\documentclass[runningheads]{llncs}
\usepackage[T1]{fontenc}
\usepackage{graphicx}
\usepackage[breaklinks]{hyperref}
\usepackage{wrapfig}

\usepackage{color}

\begin{document}
\title{Te Ahorré Un Click: A Revised Definition of Clickbait and Detection in Spanish News}
\titlerunning{Te Ahorré Un Click}
%
\author{Gabriel Mordecki\inst{1} \and
Guillermo Moncecchi\inst{1} \and
Javier Couto\inst{2}}
\authorrunning{G. Mordecki et al.}
%
\institute{Universidad de la República, Montevideo, Uruguay
\email{\{gabriel.mordecki,gmonce\}@fing.edu.uy}\\
\and PEDECIBA, Uruguay\\
\email{javier.couto@gmail.com}}
\maketitle              
\begin{abstract}
We revise the definition of clickbait, which lacks current consensus, and argue that the creation of a curiosity gap is the key concept that distinguishes clickbait from other related phenomena such as sensationalism and headlines that do not deliver what they promise or diverge from the article. Therefore, we propose a new definition: \textit{clickbait is a technique for generating headlines and teasers that deliberately omit part of the information with the goal of raising the readers' curiosity, capturing their attention and enticing them to click.} We introduce a new approach to clickbait detection datasets creation, by refining the concept limits and annotations criteria, minimizing the subjectivity in the decision as much as possible. Following it, we created and release TA1C (for \textit{Te Ahorré Un Click}, Spanish for \textit{Saved You A Click}), the first open source dataset for clickbait detection in Spanish. It consists of 3,500 tweets coming from 18 well known media sources, manually annotated and reaching a 0.825 Fleiss' $ \kappa $ inter annotator agreement. We implement strong baselines that achieve 0.84 in F1-score.

\end{abstract}

\section{Introduction}\label{sec:introduction}

Even though clickbait is a widely employed term, there remains a lack of consensus regarding its precise definition. Despite the different criteria about what clickbait is, there is some common ground: it differs from traditional headlines in their objective and style, leaving behind the goal of informing in favor of attracting attention \cite{palau-metamorphosis,cibermedios-28-paises,clickbait-estrategia-viral,oxford}. Journalists use clickbait to lure, manipulate, bait the reader into clicking, as the name implies \cite{blom-forward-reference,clickbait-estrategia-viral}. This phenomenon is also often linked to loss of journalistic quality\cite{analisis-clickbaiting,agreggation-effect,palau-metamorphosis} and the creation of disappointment among readers; and even when it does not, it is very hard to overlook because it captures our attention by creating curiosity.

Clickbait started in low-reputation web-exclusive media that focused on political propaganda or soft-news, such as The Huffington Post, Buzzfeed and Upworthy \cite{attention-merchants}, or even bad quality advertising such as in chumboxes\footnote{\url{https://en.wikipedia.org/wiki/Chumbox}}. However, it has recently gained prominence across all types of news and media.  

\begin{figure}     
  \includegraphics[width=\textwidth,keepaspectratio]{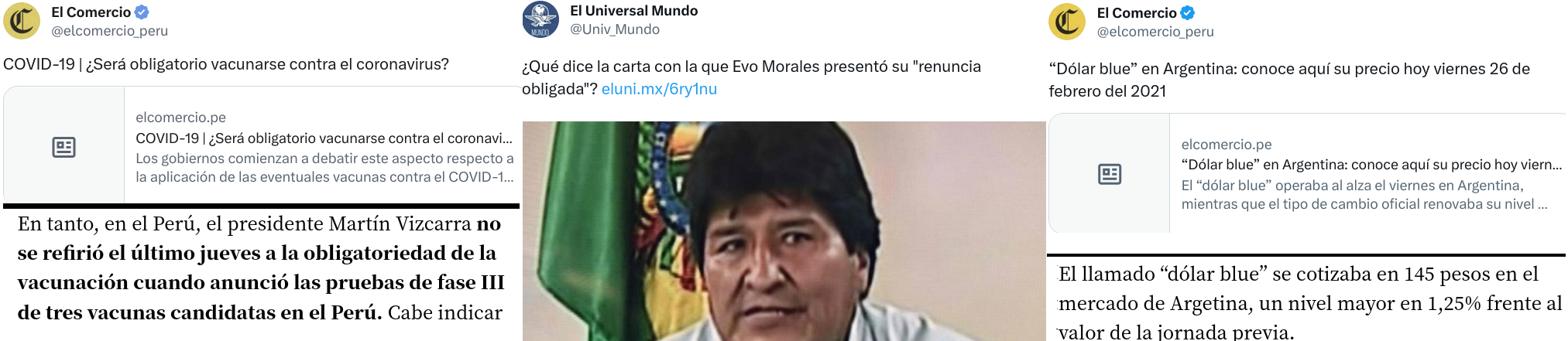}
  \caption{Three examples of clickbait news. Left: \textit{COVID-19 | Will vaccination against coronavirus be mandatory?} [in linked note:] \textit{Meanwhile, in Peru, President Martín Vizcarra did not address the mandatory nature of vaccination last Thursday [...]} An example of clickbait where it disappoints by not answering the information-gap it created. Center: \textit{What does the letter with which Evo Morales submitted his `forced resignation' say?}, clickbait on hard news. Right: \textit{Argentina's unofficial dollar: find out here its price today, Friday, February 26, 2021.} [in linked note:] \textit{The so-called `blue dollar' was trading at 145 pesos in the Argentine market [...]} A case where a clickbait news delivers exactly what it promises.}
  \label{image:tweets}
  \vspace{-0.5cm}
\end{figure}

Many authors define clickbait as a type of content \cite{oxford,clickbait-estrategia-viral} exclusively tying it to soft-news, yellow journalism and especially sensationalism, even to the point of considering them synonyms \cite{indurthi-etal-2020-predicting}. Also, several definitions make focus on that deceptive effect \cite{8-amazing-secrets,rony-diving,stop-clickbait-Chakraborty,indurthi-etal-2020-predicting} created by the news failing to deliver what they promise, a notion created by examples like the first one in \hyperref[image:tweets]{Figure~\ref{image:tweets}}.

These two attributes were essential in the early stages, but they are no longer a requirement for a headline to be clickbait, especially since it was adopted by many of the most reputable and prestigious media sources. Currently, there are plenty of cases of hard news that are clearly clickbait, such as the second example in \hyperref[image:tweets]{Figure~\ref{image:tweets}} about the words of a president overthrown by a coup d'état. Also, many teasers create curiosity and then answer exactly what they promise in the news text. Some of those news concern very concrete data like in the third news presented in the \hyperref[image:tweets]{Figure~\ref{image:tweets}}, along with the article's answer, and some answer in long articles with good journalistic quality like \textit{Africa eradicates polio: How it achieved this historic milestone.}\footnote{All of the headline and tweet examples are from the TA1C dataset, except when noted, and have been translated to English by the authors.}

All of the available clickbait detection datasets that we have knowledge of are built assuming some of these characteristics or without following a concrete definition. In \cite{potthast-etal-2018-crowdsourcing} Potthast et. al. describe three distinct methods for creating them. The first approach, named the \textit{reputation method} involves collecting teasers from media sources with established good reputations and categorizing all of them as non-clickbait whereas teasers from notorious clickbait-oriented media or websites are categorized as clickbait. The second approach is the \textit{gatekeeper method} that relies on curated third-party forums dedicated to exposing clickbait in platforms like \textit{Reddit} and \textit{Twitter}. It consists of assigning all of them to the positive class, while teasers from news-focused forums that explicitly avoid clickbait are collected as the negative class. Lastly, the \textit{importance method} involves the systematic sampling of teasers based on a predefined criteria of relevance and then manually label them.

These first two ways of creating datasets, however, lead to differences in the two classes in much more than their samples being clickbait or not. In the reputation based method, texts come from different media outlets that typically have different writing styles and do not cover the same topics. They even usually consider media that publish clickbait as part of the negative class. In a similar way, gatekeeper based methods create datasets that rely on the criteria of the selected intermediaries to choose which news to publish in their forums. Again, they typically differ in source and topics.

It is the third approach where a precise clickbait definition and annotation criteria becomes crucial. In this paper, we argue that the information gap theory \cite{Loewenstein1994} that explains why curiosity arises is not only relevant but also the fundamental difference between clickbait and other related phenomena such as sensationalism. Its essence is not about the content of the news, the language employed, or the creation of disappointment; rather, it is withholding information to provoke curiosity. We propose a new formal definition for the term in \hyperref[sec:definition]{Section~\ref{sec:definition}}.

Based on this definition, we introduce a new approach to clickbait detection datasets creation. The Clickbait Challenge corpus construction \cite{potthast-etal-2018-crowdsourcing} tackled the problem of having different perspectives of what clickbait is by relying on many annotations in a non-binary scale, their annotation task showed multiple clickbait definitions and a few examples, allowing each annotator to establish their own criteria on the boundaries of what is clickbait or not. Instead, we address these conflicting perspectives by defining clickbait as precisely as possible and refining the annotation criteria to minimize the subjectivity of the decision. Therefore, we reach a 0.825 Fleiss' $ \kappa $ inter-annotator agreement while reported agreements in the literature range from a Fleiss' $ \kappa $ of 0.36 up to 0.73 for binary annotation \cite{potthast-etal-2018-crowdsourcing,tabloids-in-the-era,Broscoteanu-EMNLP-2023,attention-grabbing}, similar to those found in other subjective NLP tasks such as irony and humor detection.

As it usually happens in NLP, most of the publications on clickbait detection work with English-written news, and as far as we know there are no publicly available datasets in Spanish. We, therefore, create and publish TA1C: a dataset for clickbait detection in Spanish and implement baseline classification methods that reach an F1-score of 0.84 and average precision of 0.93 on the positive class.

\section{Related Work}\label{sec:related-work}

Clickbait detection works are heavily influenced by dataset construction. Some influential works in the area \cite{using-dl-agrawal,stop-clickbait-Chakraborty,tabloids-in-the-era,rony-diving,8-amazing-secrets} report nearly perfect results. However, because of the way they created their datasets, machine learning models can learn to distinguish between the two classes based on features unrelated to clickbait, solving a significantly less challenging task instead.

As in the vast majority of problems in the NLP field, initial solutions were primarily traditional methods using features ranging from lexical (e.g., n-grams, word lists)\cite{potthast:2016b} and syntactic features like amount of stopwords and length of syntactic dependencies\cite{stop-clickbait-Chakraborty} to tweet-related information (e.g., attachments, posting time) \cite{potthast:2016b} to text-readability scores\cite{8-amazing-secrets}, forward-reference (a feature based in the seminal work by Blom and Hansen \cite{blom-forward-reference}) and distances between headline and article. Then, works transitioned to the prevalence of deep learning methods \cite{using-dl-agrawal,using-nn} and then to transformers architectures \cite{10.1007/978-3-030-63031-7_31,zhou2017clickbait}.

The Webis Clickbait Corpus, presented in \cite{potthast-etal-2018-crowdsourcing} is the most important dataset on clickbait detection up to date. Many works use it, with the advantage that their results can be compared to each other. The best result was obtained by fine-tuning a RoBERTa model, reaching 0.74 of F1-Score in the positive class \cite{indurthi-etal-2020-predicting}.

There are many works describing and analizing clickbait in Spanish in a qualitative way \cite{palau-metamorphosis,clickbait-estrategia-viral,analisis-clickbaiting,uso-medios-espanoles,cibermedios-28-paises}. However, there are, to the best of our knowledge, no publicly available datasets in Spanish.

\section{A clickbait definition}\label{sec:definition}

There is a general agreement that clickbait is a headline and/or teaser generation strategy that involves some sort of manipulation to bait the reader into clicking to easily increase traffic. Disagreements arise when it comes to determining which of the many methods that can be applied to increase the traffic are categorized as clickbait. It usually goes along with sensationalism because they are both answers to a usual question in the media: \textit{How to easily increase traffic?}; however, they are two different responses to it. Unlike sensationalism, the core of the clickbait concept goes beyond what the news is about or what the article says: it only concerns how the news stories are presented.

We argue that the key feature that distinguishes clickbait from other forms of attracting attention is the explicit creation of a curiosity gap. As Loewenstein describes, curiosity is in one of its interpretations, <<a form of cognitively induced deprivation that arises from the perception of a gap in knowledge or understanding>> \cite{Loewenstein1994}. It is more intense when we know some about a topic, but not all about it \cite{Kang2009}, exactly what clickbait intends to generate. When curiosity arises, it has effects in our neurocognitive mechanisms, just as food-seeking or money does, creating a strong urge to act \cite{naive-curiosity}, even involuntarily \cite{attention-grabbing}. It can lead us to irrationally seek information even when it is non-instrumental and non-relevant or even harmful, especially when we can get the answers immediately and with low cost \cite{Kang2009,KRUGER20091173,Shin2019}.

Clickbait exploits these reward systems, making us susceptible to falling into an indulgent consumption of some news \cite{SCOTT202153}. It can lead us to divert scarce resources such as time and attention to topics that don't actually interest us \cite{Shin2019}. That is why it is so effective for the media to succeed in the attention economy, and generally so annoying for the general public.

The creation of a curiosity gap intended to attract attention is the key concept to distinguish clickbait from other headlines. Therefore, we propose a definition:
\begin{quote}
  Clickbait is a technique for generating headlines and teasers that deliberately omit part of the information with the goal of raising the readers' curiosity, capturing their attention and enticing them to click.
\end{quote}

Many works on clickbait cite the curiosity gap theory and some even agree that it is what makes clickbait work \cite{attention-grabbing,SCOTT202153,curiosity-modeled,demand-avoidance-information}. However, none of them creates a dataset using this definition.

\subsection{Annotation criteria}

The proposed definition refers to the creation of an information gap as being deliberate. Since we cannot discern the author's intent, we must use an operational definition of clickbait to label the dataset: \textit{teasers that omit part of the information not obvious by context, arising curiosity and enticing users to click}.

Even with this definition, some decisions are still left to individual interpretation. In the next paragraphs we describe some criteria, with the goal of better defining clickbait and reducing as much as possible the subjectivity in the annotation.

\subsubsection{Take all the context of the news publication into account}

With this operational definition the problem of inferring the purpose of the writer disappears. However, a new one emerges: what is considered as \textit{obvious by context}?

Given the limited space available to headlines, it is a common practice to omit obvious context and even words\footnote{It has even resulted in a specific jargon called \href{https://en.wikipedia.org/wiki/Headline\#Headlinese}{Headlinese}}. In a headline like \textit{The massacre at Kabul airport results in at least 170 deaths},
and specially on Aug 27th, 2021, it is implied that readers will know what massacre they are talking about; the news is the update on casualties, not the event itself.

The boundaries between what is obvious and what is not can sometimes be blurry and subjective. The proposed criteria is to assume that anything within the common knowledge of an average reader of that publication is known, considering factors such as where, to whom, and when the news was released. In the previous example, the publication date is key, because the \textit{the massacre at Kabul airport} took place the day before. When reading teasers from foreign media, the annotator must keep in mind that the target audience of the news will be familiar with the context, and assume it as known. Annotators can even seek additional information while tagging.

The headline \textit{USA: A Donald Trump family scandal casts a shadow over the beginning of the Republican convention}
is clearly clickbait: one wonders what the scandal is about. But in very similar case of \textit{New York Governor Andrew Cuomo resigns amid the sexual harassment scandal} although it could be interpreted in the same way at first glance, it is actually referring to a widely known matter at the time. None of the targeted public at the time wondered what scandal they were referring to, so in this case, it should not be considered clickbait.

There are instances when we require subtle contextual information to correctly judge if the omitted information is not so relevant in a news article and can be excluded for the sake of brevity, or it is so crucial that its absence would result in the creation of an information gap. The case of reports about professional athletes infected with COVID illustrates the point. In \textit{Mexican golfer María Fassi tested positive for COVID-19}
it is clear that the news is about who contracted it, especially knowing that she is a mexican athlete and it comes from mexican media. Contrarily, in \textit{Juventud de las Piedras reports 17 COVID-19 positive cases}
almost all of the team is sick, we should not expect the headline to list all of them, the relevant information is the number 17. These are clear cases of non-clickbait. In \textit{A golfer tested positive for coronavirus during the PGA Tour tournament}
some context is needed to judge: the tweet is from June 2020, it was the start of the pandemic and this represented the first case. We consider that the positive case is the news by itself, so it is not clickbait even if the name of the golfer is missing. Although \textit{A Real Madrid player tested positive for coronavirus ahead of the match against Inter}
seems the same, this news is from November, matches were played anyway, so the relevant part in this case is which Real Madrid's player will not be able to play: it is clickbait. Finally, in \textit{NBA game postponed due to a positive COVID-19 case}
whether it is news by itself or whether it is relevant to which player or game, really depends on the reader's interest in the NBA. As it is such a subjective case but the news could make sense by itself, the criteria is to tag it as non-clickbait.

\subsubsection{Sensationalism is not necessarily clickbait}

Clickbait is commonly mixed up with sensationalism, but it is not the same. Some news that could be labeled as clickbait in some other works, such as the headline \textit{Leticia Brédice posted strong messages against her boyfriend: she accused him of being a `thief' and `bad person'}
are tagged as negative in this dataset: there is no information gap.

In a similar fashion, if the news itself is what generates the curiosity rather than the presented information nor the writing style, it must not be considered clickbait. It is the case of \textit{Video: Crocodile ate two sharks on an australian beach}
where the headline details exactly what is found in the news, without omitting any crucial information.

\subsubsection{Intense adjectivation may lead to clickbait}
There is a fine line when it comes to adjectivation: sometimes an excessive use of qualifications or an exaggeration is what creates the curiosity. The headline \textit{How a space hurricane is, the spectacular phenomenon detected for the first time on Earth}
is clickbait because it creates the information gap by not explaining at all why it is \textit{spectacular}. It is very common in chronicles such as \textit{Anna Delvey, the fake heiress who deceived New York's high society and whose luxurious lifestyle left substantial unpaid debts.}
Here, the headline does not exactly omit information, it highlights specific details without the needed context to understand them, creating the curiosity to fill that new information gap. On the other hand, some strong adjectives can be understood in a context, like in \textit{The bloody battle for the Sinaloa Cartel: Mayo, El Chapo's sons, and Caro Quintero's group compete for its control}
where it is well known that \textit{bloody} is not exaggerated in drug trafficking wars, therefore it is not clickbait.

\subsubsection{Self-responded gaps are not clickbait}

Frequently, newspapers tweet the same news many times with different texts describing the same article. In some cases those texts answer the uncertainty created in the headline. They are negative because they close the information gap without the necessity of clicking.

\subsubsection{Not all questions are clickbait}
Framing headlines as questions is a typical way of creating it, but the presence of one does not always imply they are clickbait. There are examples such as when the question is already answered in the teaser, it is a cited question, it is addressed directly to the reader, or when it is clear by context that it will not be answered in the article like in \textit{Football or rugby? Gaelic football, the fusion of two sports where Argentina excels}.

\subsubsection{Some direct addresses to the reader create clickbait}

Talking directly to the reader is another common way to attract its attention. Many works include it as a characteristic of clickbait \cite{palau-metamorphosis,clickbait-estrategia-viral,uso-medios-espanoles,potthast:2016b}. Sometimes, it is indeed what creates an information gap, like in the case of \textit{WhatsApp will stop working on several iPhone models: Will you need to switch to a new phone?}
However, just the direct address to the reader does not imply clickbait, in \textit{Over 15,000 Peruvians marched to reject the interim government. `This Congress does not represent me' was one of the slogans of those who protested. We inform you}, that last appeal to the reader does not create an information gap because it does not add or promise any specific information at all. It is also the case of the question in \textit{Do you remember what you were doing, where were you when Milton Wynants won the silver medal in Sydney 2000?}\footnote{Not in the dataset: \url{https://twitter.com/ovacionuy/status/1307704183928377352}} where it is obvious that it can not be answered in the article.

\subsubsection{Editorials can be clickbait}
There are works that exclude opinion articles because it is not possible to assess their truthfulness\cite{analisis-clickbaiting}. However, there are cases like \textit{Why is the Government once again targeting the peace process?}
where an editorial does omit information and arouses curiosity, they are clickbait.

\section{The TA1C dataset}\label{sec:dataset-methodology}

\begin{wrapfigure}{R}{0.5\textwidth}
  \vspace{-1cm}
  \begin{center}
    \includegraphics[width=.48\textwidth,keepaspectratio]{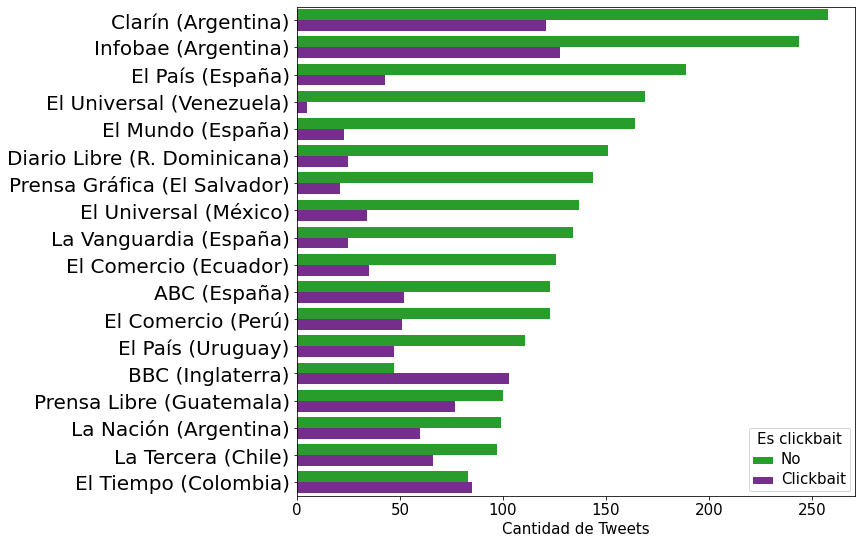}
    \caption{Tweets in the TA1C dataset by media, divided by those labeled as Clickbait and not, sorted by amount of total tweets.}
    \label{image:dataset_newspaper}
  \end{center}
  \vspace{-0.7cm}
\end{wrapfigure}

Based on the previous definition and annotation guidelines, we created the TA1C dataset: a clickbait detection corpus representative of all Spanish-language mainstream media news, available at \url{https://github.com/gmordecki/TA1C}. It is composed of news collected from Twitter, where we prioritized reputable, national or international (not local), general (not focused on one topic) and popular\footnote{Based on Twitter followers, a Comscore ranking accessed in \href{https://www.totalmedios.com/nota/41878/en-marzo-infobae-fue-el-sitio-de-habla-hispana-mas-visto-del-mundo}{the press} and the rank by \href{https://www.4imn.com/topLatin-America/}{4 International Media \& Newspapers}.} media. Also, with the objective of representing all of the Spanish language and its diversity, we chose newspapers from as many Spanish-speaking countries as possible, totaling 12, as well as international media that reach all of them. The complete list is available in \hyperref[image:dataset_newspaper]{Figure~\ref{image:dataset_newspaper}}, along with the proportion of clickbait we found in each.

\subsection{TA1C Dataset description}\label{subsec:dataset-description}

We downloaded a total of more than half a million tweets throughout a year between October 2020 and October 2021. Each tweet data was downloaded and is shared with the URL and clean HTML of the linked news, alongside the scraped headline, subheadline, clean text (only the news body), images with their captions and embedded URLs (usually social media links).

All retweets were discarded along with tweets without links or with links to other than the media main site and all of the cases where the news content crawling failed. For each newspaper we considered their main accounts and also secondary ones such as section-specific accounts. Some of them are specialized in soft-news and are more clickbait inclined, as Palau-Sampio describes \cite{palau-metamorphosis}.

The TA1C dataset contains 3500 randomly sampled tweets from those, guaranteeing that each of the 18 media sources contributes at least 150. The 3500 were annotated and split into a train set of 2100, and validation and test sets of 700 each. The splits were randomly made maintaining the same media and label balance as the complete dataset.

\subsection{Annotation methodology}

Each tweet was labeled by three independent annotators: two of the authors plus one annotator with expertise in Communication Sciences. Initially, we labeled a sample of 100 news, then deliberated on the instances with differing annotations to ensure that they were distinctions of subjectivity rather than criteria discrepancies. We reached a Fleiss' $ \kappa $ inter annotator agreement of 0.825. The final annotation is decided by majority vote, but we publish all of the annotations.

\section{Experiments and Results}\label{sec:results}

We present some baselines on clickbait detection in Spanish using the TA1C datasets. All of the reported experiments used the training set to train and the validation set to define the best model, e.g. base model for transfer learning and hyper-parameters. Reported results are from a prediction in the test set.

The text from both the tweet and the news headline were used for training purposes. We did not employ the article's text, any images, or tweet data such as date and media. They are, although, part of the dataset and could be valuable for the task.

All of the experiments run on a preprocessed text where we substitute emojis, hashtags and account at signs for special tokens and remove URLs, special characters and multiple spaces. Then, we combined the headline with the text, keeping the longer one if one includes the other, or using the headline followed by the tweet if they are distinct.

\subsection{Methods}

\textbf{LLMs - GPT4} LLMs classification with OpenAI's GPT-4 model using a \textit{You are a helpful assistant. You are designed to understand text in English and Spanish and answer in English. Your answer is just one word and one number: yes or no, followed by the probability that the answer is yes} system prompt and a \textit{Is the headline ``[Headline]'' clickbait? Answer:} user prompt. We report the GPT-4 metrics since it obtained better results than Cohere AI's LLMs.

\textbf{Has Question} True if there is a question mark in the text.

\textbf{Tf-idf + XGboost} Text vectorization with 1-3 grams TF-IDF features followed by an XGBoost classifier. We report the XGboost since it obtained better results than a linear model like LR or SVM.

\textbf{fastText} Classification with the fastText library \cite{fastText} using the supervised text version, with hyperparameter tuning via its \textit{autotuneValidation} method.

\textbf{Crafted features + XGBoost/LR} Mannualy crafted features extracted from the texts, adapted from the literature review (\cite{palau-metamorphosis,stop-clickbait-Chakraborty,uso-medios-espanoles,8-amazing-secrets,detecting-heres-how,language-independent,effective-headlines,cao2017machine}) many of them including a translation process. Both XGBoost and Logistic Regression are reported since they got the best results among other trees and linear models respectively.

\textbf{Finetuned Beto} Finetuning of Beto \cite{beto}, a language model with BERT-base architecture trained in spanish text. Up to 10 epochs of train, best hyper-parameters: batch\_size 32, dropout 0.2 and learning rate 6e-5.

\subsection{Results}

The results are available in \hyperref[table:results]{Table~\ref{table:results}}. LLM's poor performance may reflect the inherent complexity of the problem, as well as the difficulty  reconciling the different criteria for clickbait that certainly appear in their training texts. Tf-idf's and fastText bad results confirm that lexical information is not enough to discriminate clickbait. The Finetuned Beto results confirm the effectiveness of the transfer learning techniques for natural language detection problems, particularly when employed with a consistent dataset.

\begin{table*}[ht]
\centering
\resizebox{\textwidth}{!}{%
\begin{tabular}{llllll}
\hline
  \textbf{Model} & \textbf{Avg Precision} & \textbf{F1-score} & \textbf{Precision} & \textbf{Recall} & \textbf{Accuracy} \\
  \hline
  LLMs - GPT4 & 0.49 & 0.37 & 0.71 & 0.25 & 0.75 \\
  Has Question & 0.27 & 0.41 & 0.93 & 0.26 & 0.78 \\
  Tf-idf + XGboost & 0.74 & 0.61 & 0.85 & 0.48 & 0.82 \\
  fastText & 0.76 & 0.64 & 0.77 & 0.55 & 0.82 \\
  Crafted features + LR & 0.78 & 0.64 & 0.80 & 0.53 & 0.82 \\
  Crafted features + XGBoost & 0.79 & 0.69 & \textbf{0.81} & 0.60 & 0.84 \\
  Finetuned Beto & \textbf{0.90} & \textbf{0.84} & 0.80 & \textbf{0.89} & \textbf{0.90} \\
  \hline

\end{tabular}}
\caption{Results obtained in clickbait detection on the TA1C dataset. Average Precision, F1-score, Precision and Recall are reported over the positive class.}
\label{table:results}
\end{table*}

\section{Conclusion}\label{sec:conclusion}

Towards the goal of automatically detecting clickbait, we proposed a new definition of clickbait that differentiates it from other concepts like sensationalism and better fits the phenomenon after being adapted by mainstream and traditional media. We also stated precise criteria to determine whether a teaser qualifies as clickbait, in order to minimize subjectivity as much as possible.

We presented and released TA1C: the first publicly available dataset for clickbait detection in Spanish. It consists of 3,500 tweets from 18 media outlets representative of all of the Spanish language. It was manually annotated following the described criteria which helped us improve annotation agreement. We also implemented strong baselines on it. We hope that the proposed definition and criteria and the TA1C dataset can contribute to the progress of automatic clickbait classification and language resources in Spanish.

\begin{credits}

\subsubsection{\discintname}
The authors have no competing interests to declare that are
relevant to the content of this article.
\end{credits}
%
%
%
\bibliographystyle{splncs04}
\bibliography{ta1c_iberamia}

\begin{thebibliography}{10}
\providecommand{\url}[1]{\texttt{#1}}
\providecommand{\urlprefix}{URL }
\providecommand{\doi}[1]{https://doi.org/#1}

\bibitem{using-dl-agrawal}
Agrawal, A.: Clickbait detection using deep learning. In: 2016 2nd
  International Conference on Next Generation Computing Technologies (NGCT).
  pp. 268--272 (2016)

\bibitem{uso-medios-espanoles}
Araujo, A.B., Serrano-Puche, J., Jaso, M.F.N.: Uso del clickbait en los medios
  nativos digitales espa{\~{n}}oles. un an{\'{a}}lisis de el confidencial, el
  espa{\~{n}}ol, eldiario.es y ok diario. Dígitos. Revista de Comunicación
  Digital  \textbf{1}(7), ~185 (Jun 2021)

\bibitem{naive-curiosity}
Aslan, S., Fastrich, G., Donnellan, E., Jones, D.J.W., Murayama, K.: People’s
  naïve belief about curiosity and interest: A qualitative study. PLOS ONE
  \textbf{16}(9),  1--20 (09 2021)

\bibitem{clickbait-estrategia-viral}
Bazaco, {\'{A}}., Redondo, M., Sánchez-García, P.: El clickbait, como
  estrategia del periodismo viral: concepto y metodología. Revista Latina de
  Comunicación Social (74),  94–115 (ene 2019)

\bibitem{8-amazing-secrets}
Biyani, P., Tsioutsiouliklis, K., Blackmer, J.: "8 amazing secrets for getting
  more clicks": Detecting clickbaits in news streams using article informality.
  In: Proceedings of the Thirtieth AAAI Conference on Artificial Intelligence.
  p. 94–100. AAAI'16, AAAI Press (2016)

\bibitem{blom-forward-reference}
Blom, J.N., Hansen, K.R.: Click bait: Forward-reference as lure in online news
  headlines. Journal of Pragmatics  \textbf{76},  87--100 (2015)

\bibitem{detecting-heres-how}
Brogly, C., Rubin, V.L.: Detecting clickbait: Here's how to do it. Canadian
  Journal of Information and Library Science  \textbf{42}(3),  154--175 (2018)

\bibitem{Broscoteanu-EMNLP-2023}
Brosco\c{t}eanu, D.M., Ionescu, R.T.: {A Novel Contrastive Learning Method for
  Clickbait Detection on RoCliCo: A Romanian Clickbait Corpus of News
  Articles}. In: Proceedings of the 2023 Conference on Empirical Methods in
  Natural Language Processing (EMNLP). ACL (2023)

\bibitem{cao2017machine}
Cao, X., Le, T., Jason, Zhang: Machine learning based detection of clickbait
  posts in social media (2017)

\bibitem{beto}
Cañete, J., Chaperon, G., Fuentes, R., Ho, J.H., Kang, H., Pérez, J.: Spanish
  pre-trained bert model and evaluation data. In: PML4DC at ICLR 2020 (2020)

\bibitem{stop-clickbait-Chakraborty}
Chakraborty, A., Paranjape, B., Kakarla, S., Ganguly, N.: Stop clickbait:
  Detecting and preventing clickbaits in online news media. In: 2016 IEEE/ACM
  International Conference on Advances in Social Networks Analysis and Mining
  (ASONAM). pp. 9--16 (2016)

\bibitem{tabloids-in-the-era}
Chakraborty, A., Sarkar, R., Mrigen, A., Ganguly, N.: Tabloids in the era of
  social media? understanding the production and consumption of clickbaits in
  twitter. Proc. ACM Hum.-Comput. Interact.  \textbf{1}(CSCW) (dec 2017)

\bibitem{language-independent}
Coste, C.I., Bufnea, D., Niculescu, V.: A new language independent strategy for
  clickbait detection. In: 2020 International Conference on Software,
  Telecommunications and Computer Networks (SoftCOM). pp.~1--6 (2020)

\bibitem{cibermedios-28-paises}
García~Orosa, B., Gallur~Santorun, S., López~García, X.: El uso del
  clickbait en cibermedios de los 28 países de la unión europea. Revista
  Latina de Comunicación Social (72),  1261–1277 (nov 2017)

\bibitem{analisis-clickbaiting}
García~Serrano, J., Romero-Rodríguez, L.M., Hernando~Gómez, {\'{A}}.:
  Análisis del "clickbaiting" en los titulares de la prensa española
  contemporánea / estudio de caso: Diario "el país" en facebook. Estudios
  sobre el Mensaje Periodístico  \textbf{25}(1),  197–212 (mar 2019)

\bibitem{demand-avoidance-information}
Golman, R., Loewenstein, G., Molnar, A., Saccardo, S.: The demand for, and
  avoidance of, information. Management Science  \textbf{68}(9),  6454–6476
  (Sep 2022)

\bibitem{indurthi-etal-2020-predicting}
Indurthi, V., Syed, B., Gupta, M., Varma, V.: Predicting clickbait strength in
  online social media. In: Proceedings of the 28th International Conference on
  Computational Linguistics. pp. 4835--4846. International Committee on
  Computational Linguistics, Barcelona, Spain (Online) (Dec 2020)

\bibitem{fastText}
Joulin, A., Grave, E., Bojanowski, P., Mikolov, T.: Bag of tricks for efficient
  text classification. In: Proceedings of the 15th Conference of the {E}uropean
  Chapter of the Association for Computational Linguistics: Volume 2, Short
  Papers. pp. 427--431. Association for Computational Linguistics, Valencia,
  Spain (Apr 2017)

\bibitem{Kang2009}
Kang, M.J., Hsu, M., Krajbich, I.M., Loewenstein, G., McClure, S.M., yi~Wang,
  J.T., Camerer, C.F.: The wick in the candle of learning: Epistemic curiosity
  activates reward circuitry and enhances memory. Psychological Science
  \textbf{20}(8),  963--973 (Aug 2009), pMID: 19619181

\bibitem{KRUGER20091173}
Kruger, J., Evans, M.: The paradox of alypius and the pursuit of unwanted
  information. Journal of Experimental Social Psychology  \textbf{45}(6),
  1173--1179 (2009)

\bibitem{effective-headlines}
Kuiken, J., Schuth, A., Spitters, M., Marx, M.: Effective headlines of
  newspaper articles in a digital environment. Digital Journalism
  \textbf{5}(10),  1300--1314 (2017)

\bibitem{attention-grabbing}
Li, X., Zhou, J., Xiang, H., Cao, J.: Attention grabbing through forward
  reference: An erp study on clickbait and top news stories. International
  Journal of Human–Computer Interaction  \textbf{0}(0),  1--16 (2022)

\bibitem{Loewenstein1994}
Loewenstein, G.: The psychology of curiosity: A review and reinterpretation.
  Psychological Bulletin  \textbf{116}(1),  75--98 (Jul 1994)

\bibitem{agreggation-effect}
Molyneux, L., Coddington, M.: Aggregation, clickbait and their effect on
  perceptions of journalistic credibility and quality. Journalism Practice
  \textbf{14}(4),  429--446 (2020)

\bibitem{using-nn}
Omidvar, A., Jiang, H., An, A.: Using neural network for identifying clickbaits
  in online news media. In: Lossio-Ventura, J.A., Mu{\~{n}}ante, D.,
  Alatrista-Salas, H. (eds.) Information Management and Big Data. pp. 220--232.
  Springer International Publishing, Cham (2019)

\bibitem{oxford}
{Oxford English Dictionary}: Clickbait definition (2023),
  \url{https://www.oed.com/search/dictionary/?scope=Entries&q=clickbait}, Last
  accessed on 2023-08-26

\bibitem{palau-metamorphosis}
Palau-Sampio, D.: Reference press metamorphosis in the digital context:
  clickbait and tabloid strategies in elpais.com. vol.~29, pp. 63--79.
  Pamplona: Ediciones Universidad de Navarra. (2016)

\bibitem{potthast-etal-2018-crowdsourcing}
Potthast, M., Gollub, T., Komlossy, K., Schuster, S., Wiegmann, M., {Garces
  Fernandez}, E., Hagen, M., Stein, B.: {Crowdsourcing a Large Corpus of
  Clickbait on Twitter}. In: Bender, E., Derczynski, L., Isabelle, P. (eds.)
  27th International Conference on Computational Linguistics (COLING 2018). pp.
  1498--1507. The COLING 2018 Organizing Committee (Aug 2018)

\bibitem{potthast:2016b}
Potthast, M., K{\"o}psel, S., Stein, B., Hagen, M.: {Clickbait Detection}. In:
  Ferro, N., Crestani, F., Moens, M.F., Mothe, J., Silvestri, F., {Di Nunzio},
  G., Hauff, C., Silvello, G. (eds.) Advances in Information Retrieval. 38th
  European Conference on IR Research (ECIR 2016). Lecture Notes in Computer
  Science, vol.~9626, pp. 810--817. Springer, Berlin Heidelberg New York (Mar
  2016)

\bibitem{rony-diving}
Rony, M.M.U., Hassan, N., Yousuf, M.: Diving deep into clickbaits: Who use them
  to what extents in which topics with what effects? In: Proceedings of the
  2017 IEEE/ACM International Conference on Advances in Social Networks
  Analysis and Mining 2017. p. 232–239. ASONAM '17, Association for Computing
  Machinery, New York, NY, USA (2017)

\bibitem{SCOTT202153}
Scott, K.: You won't believe what's in this paper! clickbait, relevance and the
  curiosity gap. Journal of Pragmatics  \textbf{175},  53--66 (2021)

\bibitem{Shin2019}
Shin, D.D., Kim, S.i.: Homo curious: Curious or interested? Educational
  Psychology Review  \textbf{31}(4),  853--874 (Dec 2019)

\bibitem{curiosity-modeled}
Venneti, L., Alam, A.: How curiosity can be modeled for a clickbait detector
  (2018)

\bibitem{10.1007/978-3-030-63031-7_31}
Wu, C., Wu, F., Qi, T., Huang, Y.: Clickbait detection with style-aware title
  modeling and co-attention. In: Sun, M., Li, S., Zhang, Y., Liu, Y., He, S.,
  Rao, G. (eds.) Chinese Computational Linguistics. pp. 430--443. Springer
  International Publishing, Cham (2020)

\bibitem{attention-merchants}
Wu, T.: The raise of clickbait. In: The Attention Merchants, chap.~22, pp.
  276--289. Knopf (2016), iSBN 9780804170048

\bibitem{zhou2017clickbait}
Zhou, Y.: Clickbait detection in tweets using self-attentive network (2017)

\end{thebibliography}
\end{document}